\pdfoutput=1

\documentclass[12pt]{l4dc2020} 

\usepackage{algorithm}
\usepackage[noend]{algpseudocode}
\usepackage{caption}
\usepackage{stmaryrd,mathtools}
\usepackage{subcaption}
\usepackage{booktabs}
\usepackage{multirow}
\usepackage{float}
\usepackage{wrapfig}
\usepackage{times}

\title{Model-Predictive Control via Cross-Entropy and Gradient-Based Optimization}

\author{%
  Homanga Bharadhwaj$^*$, Kevin (Cheng) Xie$^*$, and Florian Shkurti\\
  \texttt{Department of Computer Science, University of Toronto\\University of Toronto Robotics Institute\\Vector Institute\\} 
  \\
}

\begin{document}

\maketitle
  \vspace*{-0.5cm}
\begin{abstract}
   Recent works in high-dimensional model-predictive control and model-based reinforcement learning with learned dynamics and reward models have resorted to population-based optimization methods, such as the Cross-Entropy Method (CEM), for planning a sequence of actions. To decide on an action to take, CEM conducts a search for the action sequence with the highest return according to the dynamics model and reward. Action sequences are typically randomly sampled from an unconditional Gaussian distribution and evaluated on the environment. This distribution is iteratively updated towards action sequences with higher returns. However, this planning method can be very inefficient, especially for high-dimensional action spaces. An alternative line of approaches optimize action sequences directly via gradient descent, but are prone to local optima. We propose a method to solve this planning problem by interleaving CEM and gradient descent steps in optimizing the action sequence. Our experiments show faster convergence of the proposed hybrid approach, even for high-dimensional action spaces, avoidance of local minima, and better or equal performance to CEM. Code accompanying the paper is available here\footnote{\url{https://github.com/homangab/gradcem} }.
  
\end{abstract}

\section{Introduction}
High-dimensional, nonlinear Model-Predictive Control (MPC) and Model-Based Reinforcement Learning (MBRL) have seen significant progress over the last years, the task being to first learn a dynamics and a reward model of the environment and then plan using the learned models. While a number of recent approaches~\cite{planet,dads,pets,poplin} have developed efficient techniques for learning these models in MBRL, fewer papers~\cite{dcem,upn} have investigated the planning problem. Instead, many state-of-the-art MBRL approaches perform planning either using the Cross-Entropy Method (CEM)~\cite{cem,pets, cross_entropy_planning}, or via Model-Predictive Path Integral (MPPI)~\cite{mppi}. Both these approaches are population-based search heuristics that sample random actions, execute them under the currently learned model, obtain the sum of rewards, and update the sampling distribution to increase the probability of higher reward action sequences. In MPC, the first action of the sequence is executed in the environment, the remaining planned actions are typically discarded, and the search procedure repeats.


Many current MBRL approaches do not leverage gradients through the model, which are cheaply available, and resort to inefficient optimization, particularly in high dimensions, whereas gradient-based planning converges faster.

In this paper we combine the two methods, to take advantage of the convergence speed of gradient-based planning and the broader search, multi-extremum optimization performed by CEM. Gradient based optimization is one of the main approaches for a number of high-dimensional non-convex optimization problems in machine learning, yet it has not been widely adopted in planning problems due to the issue of vanishing or exploding gradients. We investigate situations where gradient-based planning fails, due to the sensitivity of shooting methods and imperfect models, and provide a simple way to mitigate it: we interleave CEM steps with gradient-based optimization, so that the latter can inform the update of the sampling distribution in the former. 



\section{Preliminaries}
In this section we discuss the CEM method, and some of the key issues in gradient-based planning.
\subsection{The Cross-Entropy Method for Planning}
In model-based reinforcement learning and model predictive control, a model of the environment and reward is learned from real transitions in the environment. To select an action, MPC searches for an optimal action sequence under the learned model and executes the first action of that sequence, discarding the remaining actions.
Typically this search is repeated after every step in the real environment, to account for any prediction errors by the model and to get feedback from the environment. In many works this planning step is done using the Cross-Entropy Method (CEM)~\cite{pets,planet, poplin, cross_entropy_planning}. CEM samples action sequences from a time-evolving distribution, usually a diagonal Gaussian $a_{t:t+H}\sim\mathcal{N}(\mu_{t:t+H},\texttt{diag}(\sigma^2_{t:t+H}))$.
These open-loop action sequences are simulated using the learned dynamics model to obtain approximate resulting state sequences and rewards.  By repeatedly sampling random action trajectories, evaluating them under the model, and re-fitting the sampling distribution to the best $K$ trajectories, a new Gaussian distribution $\mu_{t:t+H},\sigma^2_{t:t+H}$ of actions for the current time-step is obtained. Convergence analysis for CEM for rare event simulation is given in~\cite{cem_convergence}.  

Sampling random action sequences in this manner and evaluating the sum of rewards from them is very costly in practice because it does not leverage any implicit structure in the planning problem, and does not take advantage of the fact that gradients through the model can in fact be used to direct the search procedure, instead of naively sampling random action sequences. 

\subsection{Gradient-Based Planning}
Gradient-based methods for planning typically correspond to backpropagating derivatives of a cumulative loss (or reward) function with respect to actions for updating the sequence of actions iteratively through gradient descent. 
In~\cite{discretecontinuous}, the gradients of the cumulative reward with respect to actions are computed by differentiating through the learned reward and forward dynamics models. In~\cite{upn}, gradients of the inner loss of the Gradient-Descent Planner (GDP) with respect to actions are computed in the latent space, by differentiating through a learned latent forward dynamics model. The ultimate aim is to update actions through an iterative gradient descent approach:

$$\bar{a}_{0:H}^{(k)} := \bar{a}_{0:H}^{(k-1)} + \alpha \nabla_{\bar{a}} \bar{R}(\bar{a}_{0:H}^{(k-1)}, \bar{s}_{0:H}^{(k-1)}), \qquad k=1,2, ...,K$$

\noindent Here, $H$ denotes the time-horizon of the episode, $a$ denotes action and $s$ denotes state. One of the most important drawbacks of gradient descent for non-convex optimization is that the optimization procedure is only guaranteed to converge to a local optima, not the global optima. In MPC for MBRL, these planners may converge to sub-optimal plans. In addition, for a long horizon $H$, there is the exploding and vanishing gradients problem which must be taken care of during optimization. An important point to note is that when the action dimension increases, CEM becomes highly inefficient and requires significantly more optimization epochs due to a blow-up of the search space, whereas there is only a slight increase (one gradient dimension) in computational burden for gradient descent. This is because, for optimization CEM utilizes just the aggregate reward which is a one-dimensional feedback signal per rollout, while gradient-based planning makes use of an $D$-dimensional feedback signal, namely the gradient of the cumulative reward with respect to the actions. 


\section{Approach}
\label{sec:method}
Our approach is based on the motivation that in MPC, model gradients should be effectively used for conducting a more informed search during the planning phase. 
In the subsequent subsections, we describe a simple technique for doing this in practice.
\subsection{CEM+Gradient Descent}
Since gradient descent is prone to getting stuck at local optima and in practice requires sufficiently different random initializations to alleviate this, we consider a very simple idea - interleave CEM steps with gradient descent on the samples to locally refine each plan. This method incorporates gradients through the model, thereby yielding more refined action sequences that can be used to update the CEM sampling distribution faster. Instead of resampling all plans, we choose to keep the top K plans from the previous iteration to continue optimizing them via gradient descent.

Let $f_\phi$ denote the learned dynamics model, $r_\psi$ the learned reward model, $a_h$ the action at time-step $h$, $s_h$ the state of the environment at time-step $h$, and $H$ the planning horizon. Let $\mathcal{N}(\mu^{(t)}_{0:H},\Sigma^{(t)}_{0:H})$ denote the CEM sampling distribution from which action sequences are sampled in the $t^{th}$ CEM iteration. 
Here our notation for $\mathcal{N}$, refers to $H$ independent multivariate Gaussian distributions. We arbitrarily set the parameters $(\mu^{(0)}_{0:H}=0,\Sigma^{(0)}_{0:H}=I)$ of this distribution initially. At the beginning of each CEM iteration, the planner first samples multiple ($G$) random action sequences:
$$\{(a^{(t)}_0,....,a^{(t)}_H)_g\}_{g=1}^G \sim  \mathcal{N}(\mu^{(t)}_{0:H},\Sigma^{(t)}_{0:H})$$
We next evaluate the cumulative reward obtained from each of these action sequences, under the current learned dynamics model $f_\phi$ and the current reward model $r_\psi$:

$$R^{(t)}_g = \sum_{h=1}^Hr_\psi(s^{(t)}_h) \qquad s^{(t)}_h=f_\phi(s^{(t)}_{h-1},a^{(t)}_{h-1}), \qquad \forall g=1,..,G$$ 

\noindent Here, $t$ indexes the CEM iterations. Now, treating these initial sampled plans as initialization of the gradient-descent procedure, we perform $J$ steps of gradient descent on all of the sequences. In all of our experiments to ensure fair comparison to CEM, we set $J=1$.
$$(a^{(t)}_0,....,a^{(t)}_H)^{j+1}_g \longleftarrow (a^{(t)}_0,....,a^{(t)}_H)^j_g - \beta\nabla_{a^{(t)}_{0:H}}R^{(t)}_g, \qquad \forall g=1,..,G, j=1,...,J$$

\noindent Then we update the parameters of our proposal (sampling) distribution $\mathcal{N}(\mu^{(t+1)}_{0:H},\Sigma^{(t+1)}_{0:H}) $ to match the top $K$ updated action sequences:

$$ \mu^{(t+1)}_{0:H} \longleftarrow \text{Mean}(\{(a^{(t)}_0,....,a^{(t)}_H)_k\}_{k=1}^K )$$

$$ \Sigma^{(t+1)}_{0:H} \longleftarrow \text{Variance}(\{(a^{(t)}_0,....,a^{(t)}_H)_k\}_{k=1}^K )$$

\noindent Finally, we replace the bottom $G-K$ action sequences, with samples from the updated proposal distribution.
After $T$ iterations of this, the remaining action sequence with the highest reward is returned. Our approach is summarized in Algorithm~\ref{alg:algorithm}.



\begin{figure}
    \centering
    \includegraphics[width=0.7\textwidth]{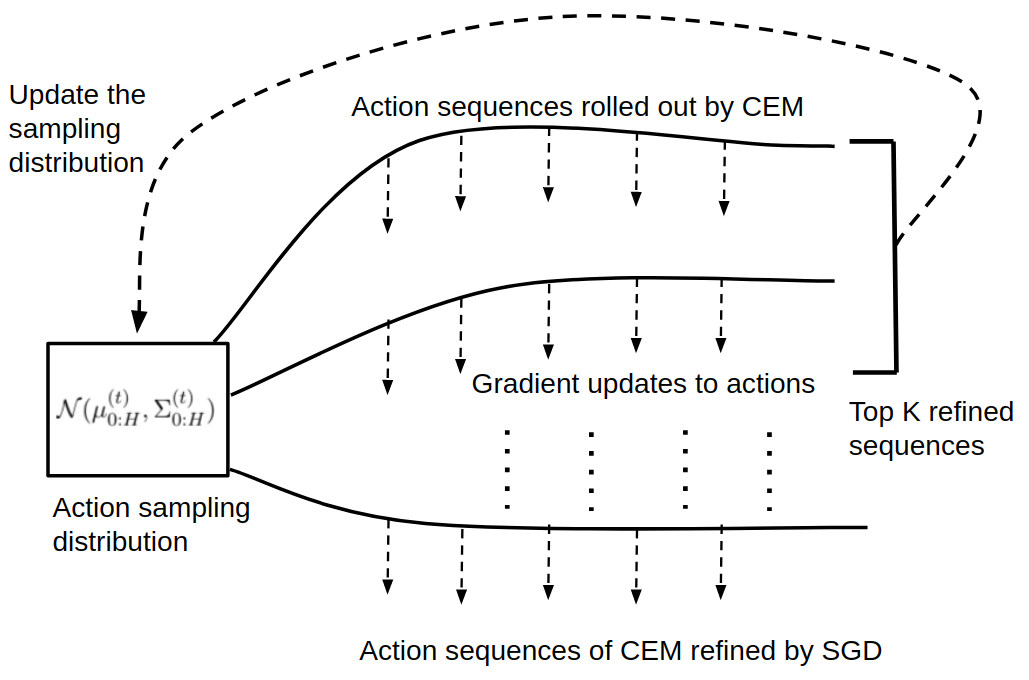}
    \caption{ Schematic of the proposed approach. Initial sequences of actions sampled from the CEM sampling distribution are refined by a few gradient descent updates, denoted by downward arrows. Then the action sequences are evaluated under the current model. The top $K$ action sequences (i.e. the top K ones with maximum sum of discounted rewards) are used to refine the CEM sampling distribution from which the actions are sampled. The sampling distribution is typically assumed to be a Gaussian, and is so for our paper as well.}
    \label{fig:schema}
       \vspace*{-0.5cm}
\end{figure}
\begin{algorithm}[h!]
    Initialize environment transitions data $\mathcal{D} \shortleftarrow \{ \}$
    
    \For{trial m=1 to M}{
        Train dynamics model $s_{1:H}=f(a_{0:H}, s_0)$, reward model $r(s_{1:H})$ with $\mathcal{D}$
        
        \For {environment step l=1 to L}{ 
        Sample $G$ initial plans $\{{a^{(0)}_{0:H}}\}_{g=1}^G$ from  $\mathcal{N}(\mu^{(0)}_{0:H}=\mathbf{0},\Sigma^{(0)}_{0:H}=I)$
        
        Sample a random environment state $s_0$
        
        \For{CEM iteration t=1 to T}{
            \For{gradient descent step j=1 to J}{
                
                $s^{(t)}_{1:H} = f(\{{{a^{(t)}_{0:H}}}\}_g,s_0)$ \quad
                
                Calculate model returns for each plan $R_g^{(t)}=r(s^{(t)}_{1:H})$ 
            
                Update $\{{a^{(t)}_{0:H}}\}_{g=1}^G$ by maximizing total model returns via SGD
            }
            
            $s^{(t)}_{1:H} = f(\{{{a^{(t)}_{0:H}}}\}_g,s_0)$ \quad
            
            Calculate model returns for each plan $R_g^{(t)}=r(s^{(t)}_{1:H})$
            
            Sort $\{{a^{(t)}_{0:H}}\}_{g=1}^G$ based on total model returns $\{R_g^{(t)}\}_{g=1}^G$ on step $J$
            
            Update $(\mu^{(t)}_{0:H},\Sigma^{(t)}_{0:H})$ to fit the top $K$ action sequences 
            
            Replace bottom $G-K$ action sequences with samples from  $\mathcal{N}(\mu^{(t)}_{0:H},\Sigma^{(t)}_{0:H})$
        }

        Execute first action from the highest model return action sequence
        
        Record real transition in $\mathcal{D}$ 
        }
    }
    \caption{Grad+CEM Algorithm (The proposed approach)}
    \label{alg:algorithm}
\end{algorithm}



\section{Experiments}
Through the experiments we aim to demonstrate the benefits and pitfalls of CEM and gradient descent, and demonstrate the efficacy of the proposed approach. The gradient-based planner baseline is hereafter referred to as Grad. This is implemented as SGD (Stochastic Gradient Descent). It samples $G$ initial samples and separately performs $T$ stochastic gradient steps on them. To better demonstrate our claims, we created a toy environment, the details of which are described in the next sub-section. Code for the experiments is available in this repository \url{https://github.com/homangab/gradcem}.

\begin{wrapfigure}{r}{0.5\textwidth}
    \centering
    \includegraphics[width=0.35\textwidth]{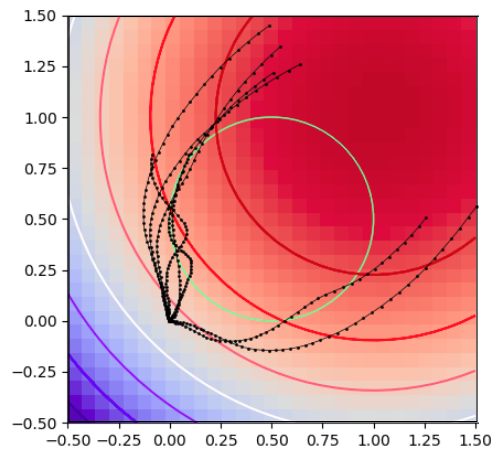}
    \caption{Illustrative diagram of the toy environment. The black paths are 2D projections of multiple paths of a point mass. Red denotes high reward and blue denotes low reward regions. The green circle is an obstacle with soft contact.}
    \label{fig:toyenv}
       \vspace*{-0.5cm}
\end{wrapfigure}

\subsection{Details of the toy environment}

To consider the planning problem in isolation, we created a toy environment in which we have access to ground truth gradients through the dynamics model. The agent controls a mass in an  N dimensional space by applying forces at each time step.
Fig.~\ref{fig:toyenv} shows a 2D projection of the environment. 
The task is to move towards high reward regions of the state-space (red region) from the blue region. 
The black lines and dots show 2D projections of multiple rolled out trajectories starting from the origin. 
The fluorescent green region denotes an obstacle with soft contact.
The soft contact is modeled as a repulsive spring force at every time step that increases proportionally to the penetration depth of the agent into the obstacle.
The ``hardness" of the contact can be tuned by the spring constant. The larger the spring constant is, the stronger is the repulsion force. For all the toy environment results, all the methods used $T=10$ number of iterations. 
To make a fair comparison we set the number of inner gradient steps per iteration $J=1$ for Grad+CEM. 
All methods used $G=20$ sampled plans at each iteration. CEM and Grad+CEM both select the top $K=4$ plans at each iteration. 

\subsection{Results in high dimensions}
\begin{figure}\begin{minipage}[b]{.49\linewidth}\centering\large \includegraphics[width=\textwidth]{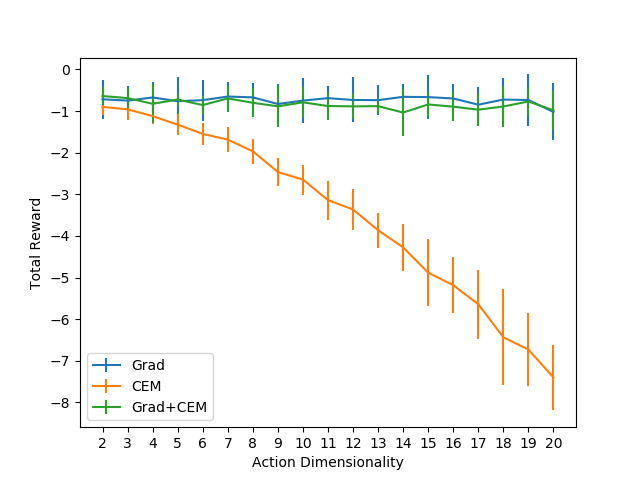}\subcaption{Performance of different planners for single obstacle, soft contact case as the environment dimensionality (both states and actions) is increased.}\label{fig:cemvsgrada}\end{minipage}%
\hfill
\begin{minipage}[b]{.49\linewidth}\centering\large  \includegraphics[width=\textwidth]{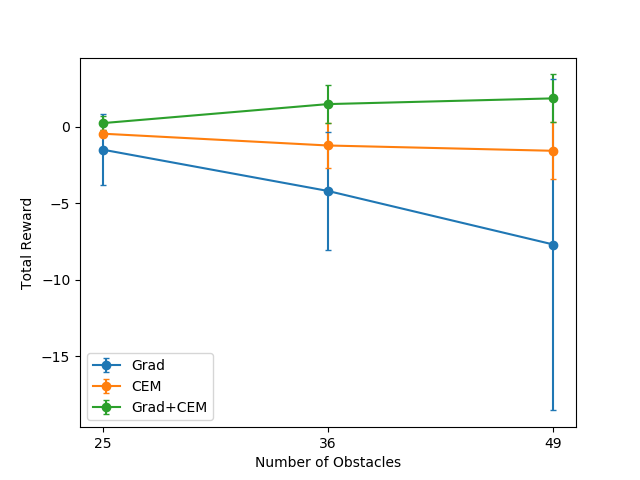}\subcaption{Multi-object, hard contact two-dimensional case as the number of obstacles is increased. For simulating hard contact, the spring constant 10 times larger.}\label{fig:cemvsgradb}\end{minipage}\caption{Total reward obtained by CEM vs Grad vs Grad+CEM planners on the toy environment. The total reward is averaged over 50 runs for each data point and error bars denote the standard deviation of those runs. The error bars correspond to one standard deviation for optimization with 50 random seeds. Higher is better. 
}
\label{fig:cemvsgrad}\end{figure}

In the toy environment shown in Fig.~\ref{fig:toyenv}, we hypothesize that increasing the dimensions of the action space is likely to deteriorate the performance of a vanilla CEM planner but not of a gradient descent based planner. Fig.~\ref{fig:cemvsgrada} shows a comparative analysis of total reward collected in the environment as the number of action dimensions are increased from 2 to 20.

There is a significant drop in the performance of the CEM based planner with increasing action dimensionality. 
For optimization CEM utilizes just the aggregate reward, which is a one-dimensional error signal per rollout, while gradient-based planning makes use of a $D$-dimensional error signal, namely the gradient. 

\subsection{When gradient based optimization fails}
\begin{figure}
    \centering
    \includegraphics[width=0.8\textwidth]{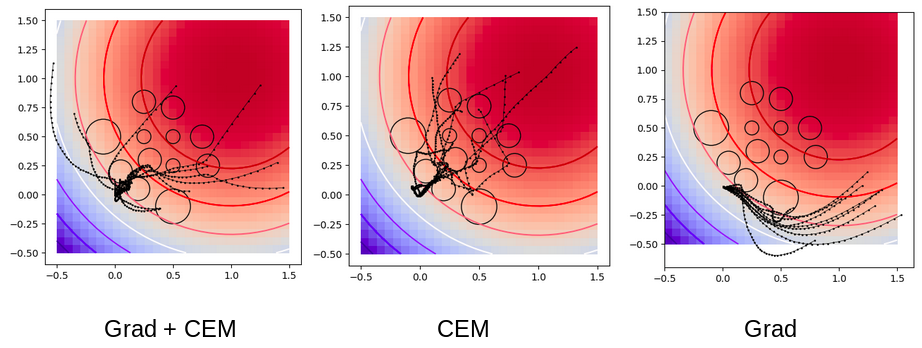}
    \caption{Illustration of trajectories of different algorithms in the multiple obstacles scenario.}
    \label{fig:gradfails}
    \vspace*{-0.5cm}
\end{figure}

Fig.~\ref{fig:gradfails} shows an experimental scenario that involves multiple obstacles with non-smooth contact (the spring constant is set 10 times higher). Here it is evident that the purely gradient based approach does not succeed and gets stuck in some local optima. The main reason for this is that the non-smooth contact results in discontinuous gradients (e.g. consider the edge of a table. 
There is a sudden jump in the magnitude of the gradients when moving from one edge to the other) which make learning difficult. To alleviate this, we show in Fig.~\ref{fig:cemvsgradb} that interleaving CEM and gradient-descent update steps helps learn better plans.
Note that we decrease the size of obstacles as we increase their number in order to pack them into the same space and that is why it is possible for Grad+CEM to do better as the number of obstacles increases.

\subsection{Experiments with Planning over Learned Dynamics Models} 
\begin{figure}\begin{minipage}[b]{.5\linewidth}\centering\large \includegraphics[width=\textwidth]{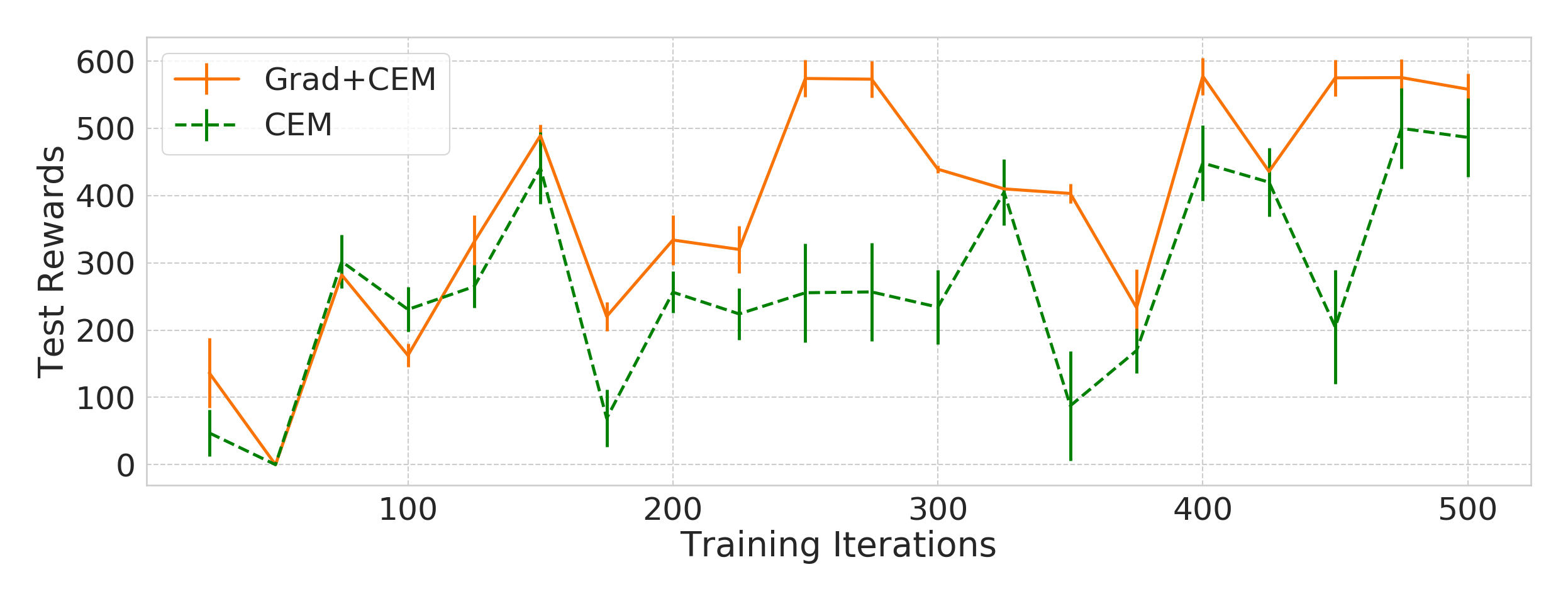}\subcaption{OpenAI Gym Half-Cheetah~\cite{gym}}\label{fig:cheetah}\end{minipage}%
\begin{minipage}[b]{.5\linewidth}\centering\large  \includegraphics[width=\textwidth]{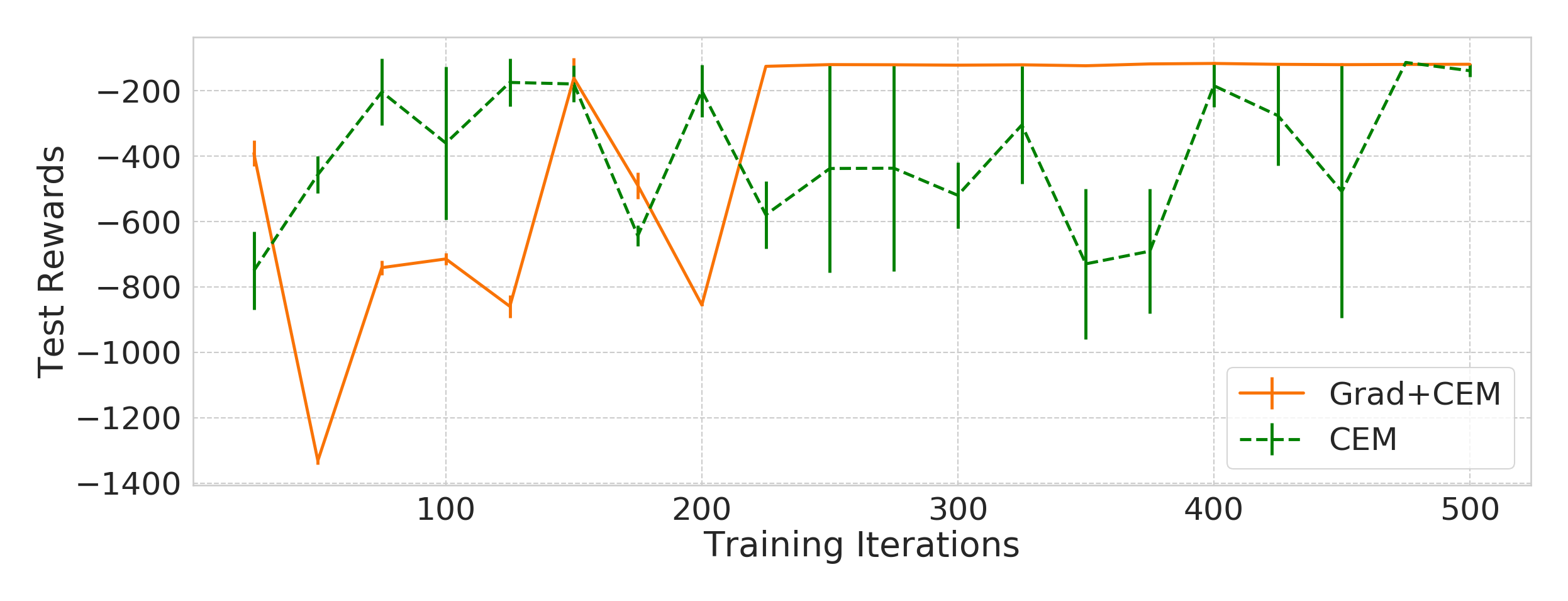}\subcaption{OpenAI Gym Pendulum~\cite{gym}}\label{fig:pendulum}\end{minipage}\caption{Variation of rewards at test time during the course of training. OpenAI Gym (a) Pendulum and (b) Half-Cheetah environments. CEM is the default Planet~\cite{planet} algorithm that plans through CEM. Grad+CEM is the version of Planet that plans using the proposed Grad+CEM scheme. The error bars correspond to the standard deviation during evaluation with three random seeds. Higher is better.}\label{fig:5}\end{figure}

In this section, we consider the complete MBRL problem of learning dynamics+reward models and using the learned models to do planning. In particular, we consider the SOTA Planet~\cite{planet} model and replace the CEM based planning module with the proposed Grad+CEM approach. Fig.~\ref{fig:5} shows that for two different OpenAI Gym~\cite{gym} environments, Pendulum and Half-Cheetah, while the default CEM based planning scheme struggles to converge in terms of the test rewards, incorporating model gradients for planning ensures a quick and reliable convergence. So, from the experiments we conclude that the Grad+CEM scheme helps in converging to higher rewards faster, with fewer optimization iterations. For Fig.~\ref{fig:pendulum} and Fig.~\ref{fig:cheetah}, for a pairwise t-test between the two variants CEM and Grad+CEM, we respectively obtain p-values 0.019 and 0.004. Both results are significant at $p<0.05$. In Fig.~\ref{fig:dmc}, conducting a pairwise t-test for (a), (b), (c), and (d), we respectively obtain p-values 0.314, 0.103, 0.121, and 0.136. These results are not significant at $p<0.05$.  In the Pendulum environment, the pendulum starts at a random position, and the goal is to swing it up so that it stays upright. In the Half-Cheetah environment, the agent gets rewarded for moving a fast as possible and maintaining proper gait (not toppling over). In both these environments, the input to the policy are high dimensional rendered images, which make the tasks challenging. Our main conclusions from these experiments are that the hybrid method has equal or better search performance compared to CEM. The main advantages of the proposed hybrid method, as shown in Figs.~\ref{fig:toyenv}, \ref{fig:cemvsgrad}, and \ref{fig:gradfails} include faster speed of convergence compared to CEM, when the dimensionality of the action space increases, as well as broader coverage of local minima than gradient-based optimization.


\begin{figure}
\centering
\begin{minipage}[b]{.49\linewidth}\centering\large \includegraphics[width=\textwidth]{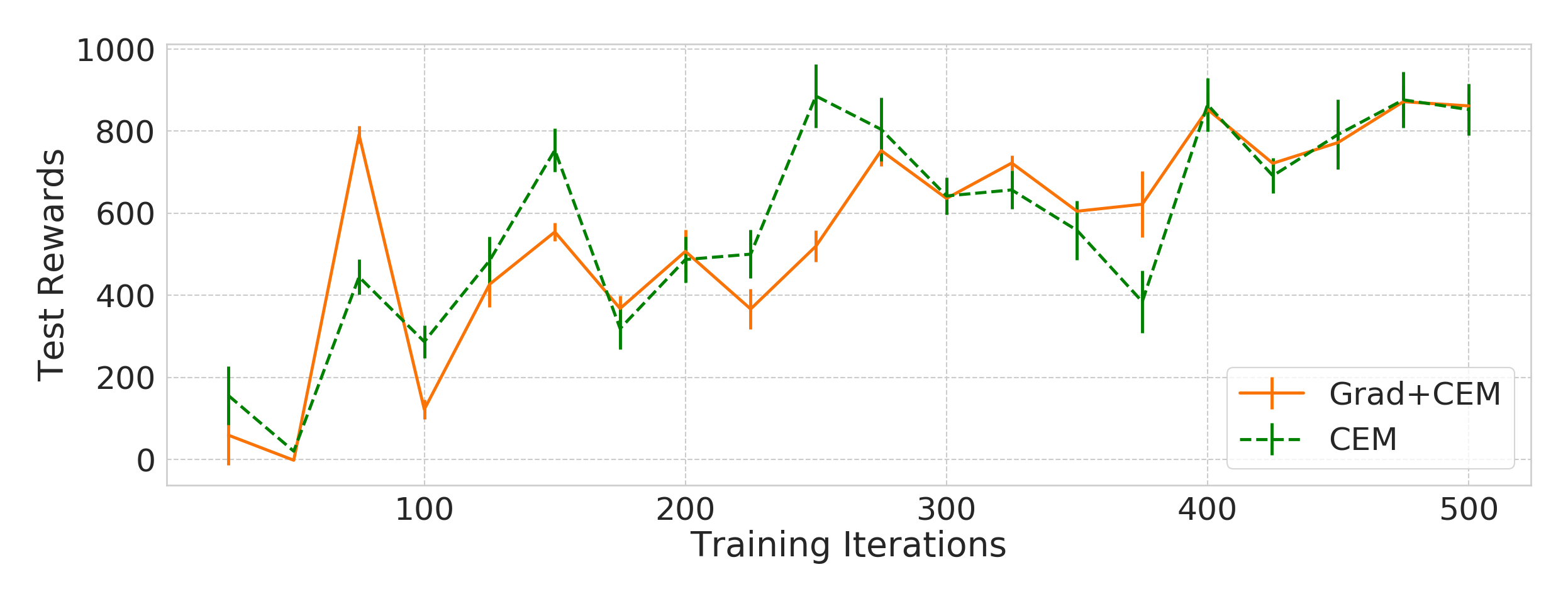}\subcaption{DM Control Suite Cartpole-SwingUp}\label{fig:2a}\end{minipage}%
\begin{minipage}[b]{.49\linewidth}\centering\large  \includegraphics[width=\textwidth]{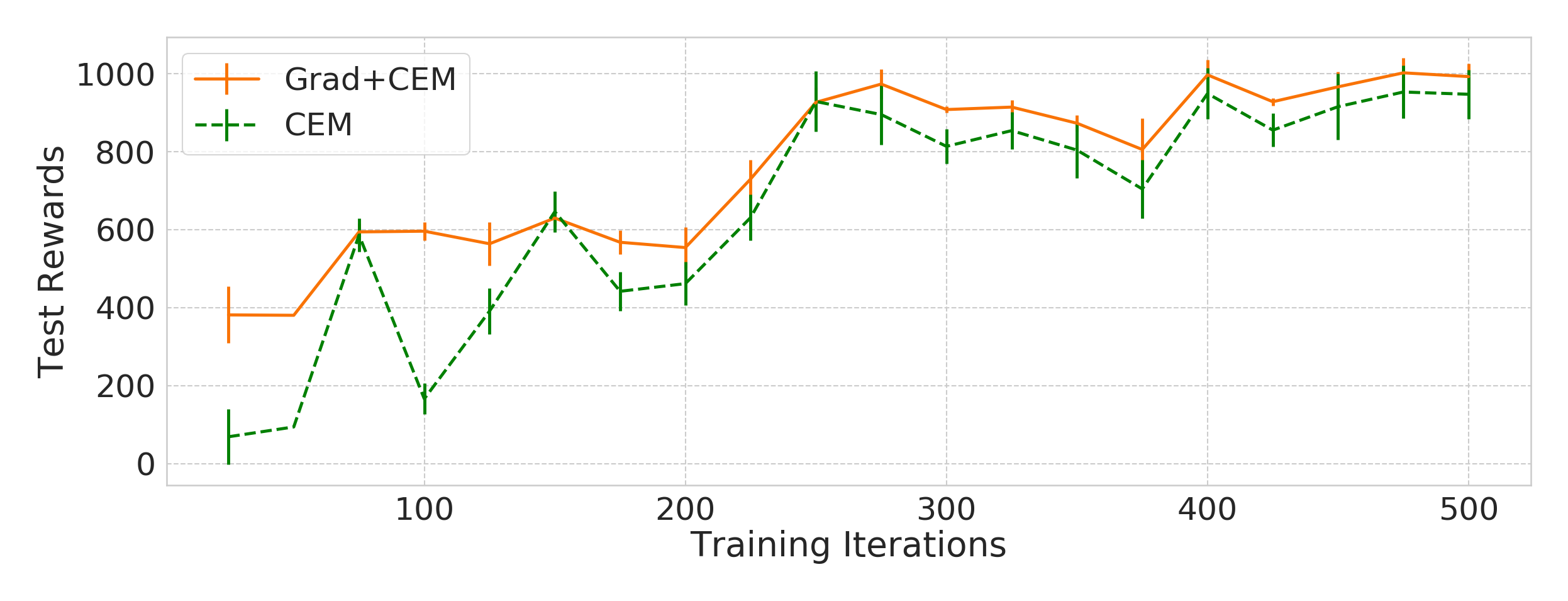}\subcaption{DM Control Suite WalkerWalk}\label{fig:2b}\end{minipage}

\begin{minipage}[b]{.49\linewidth}\centering\large  \includegraphics[width=\textwidth]{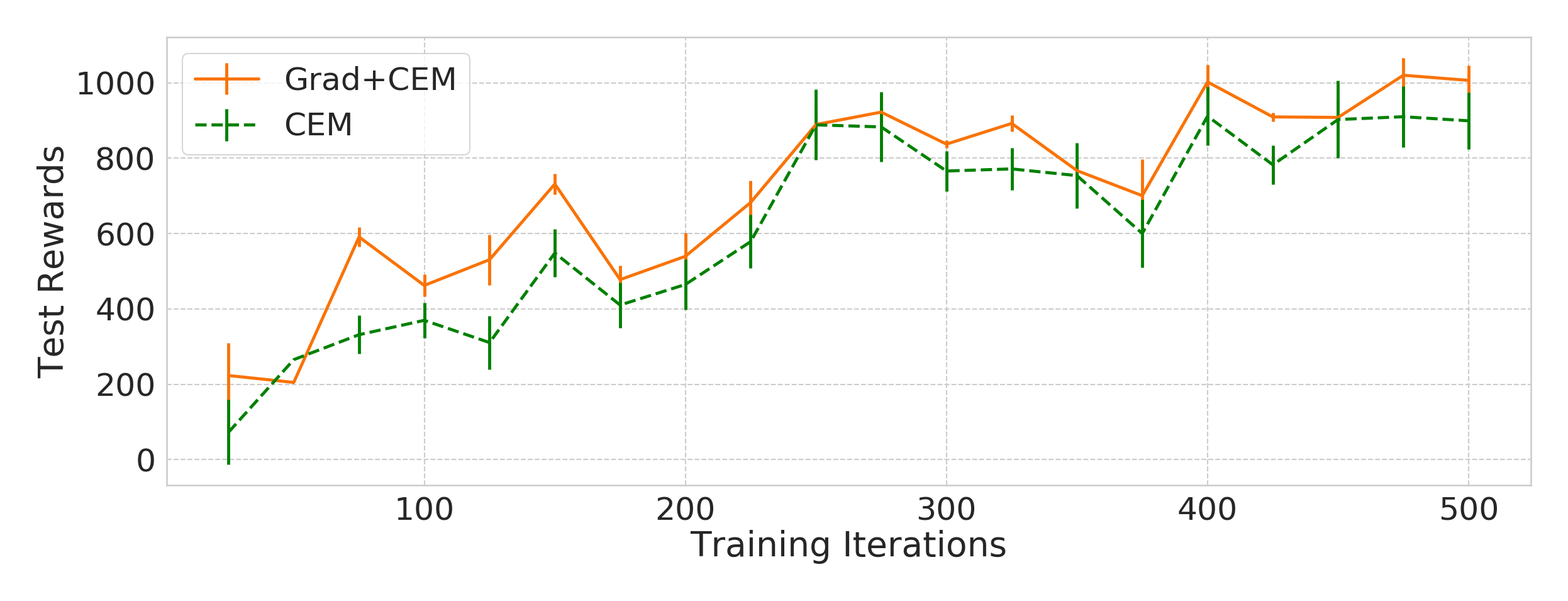}\subcaption{DM Control Suite Reacher}\label{fig:2c}\end{minipage}
\begin{minipage}[b]{.49\linewidth}\centering\large  \includegraphics[width=\textwidth]{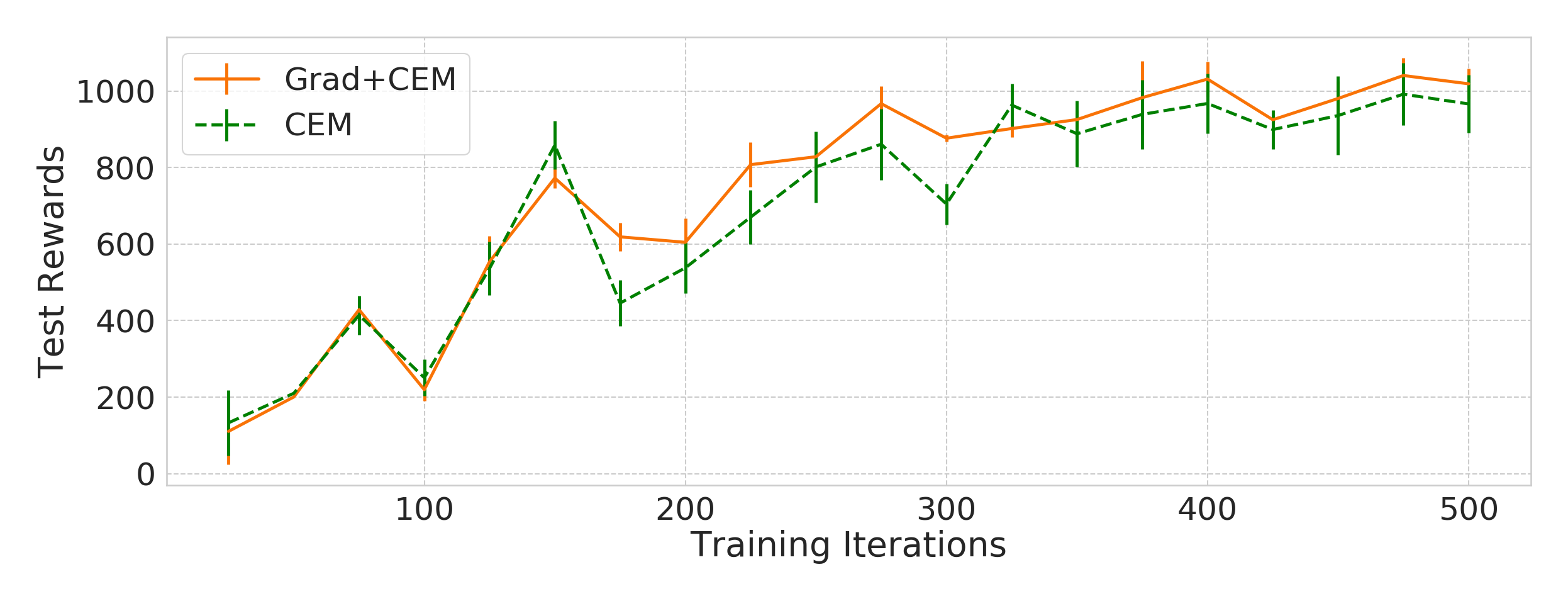}\subcaption{DM Control Suite Cartpole-Balance}\label{fig:2c}\end{minipage}
\label{fig:dmc}
\caption{Variation of rewards at test time during the course of training. DeepMind Control Suite~\cite{dmc} (a) Cartpole-SwingUp, (b) Walker-Walk,  (c) Reacher, and (d) Cartpole-Balance environments. CEM is the default Planet~\cite{planet} algorithm that plans through CEM. Grad+CEM is the version of Planet that plans using the proposed Grad+CEM scheme. The error bars correspond to standard deviation during evaluation with three random seeds. Higher is better.}\label{fig:dmc}\end{figure}

\section{Related Works} 
Our paper is broadly based on the theme of model-based reinforcement learning (MBRL)~\cite{pets,planet}, where the idea is to learn a dynamics model of the world and plan using the learned dynamics model. For high dimensional inputs like images, the dynamics are typically learned in a learnt latent space~\cite{planet}. Most current MBRL approaches use some version of the Cross-Entropy Method (CEM) for doing a random population based search of plans given the current model~\cite{poplin,planet}. Some papers~\cite{dads} use other population based search approaches like Model-Predictive Path Integral (MPPI)~\cite{mppi} for planning. A recent paper~\cite{vimpc} discusses how in the control as inference framework, the two approaches, CEM, and MPPI are very similar, and differ only with respect to the reward function $r(s_{1:H})$. Both these random shooting approaches are very costly and take a long time to converge because they involve sampling random action sequences and evaluating them under the current model to determine the high performing sequences. Although~\cite{poplin} introduces the idea of performing the CEM search in the parameter space of a distilled policy, it still is very costly and requires a lot of samples for convergence. 

Gradient-descent based optimization methods have been successful in a wide range of machine learning domains~\cite{maml}, but for planning, there are very few papers that have been able to successfully perform gradient-descent based planning. Universal Planning Networks (UPNs)~\cite{upn} optimizes action sequences in a latent space such that the optimized sequence matches expert demonstrations of actions. 
So the approach requires high quality expert data, and is based on imitation learning, not end-to-end reinforcement learning. SGD for model predictive control is also done in~\cite{discretecontinuous} but without a diverse initialization it can lead to local optima. Hence the approach is limited to simple grid worlds, and cannot scale to more challenging robotic tasks~\cite{shooting1,shooting2}.

In the context of model-free reinforcement learning, ~\cite{cem-rl} also introduce the idea of interleaving CEM and policy gradient steps in optimizing in the parameter space of policies. We show how interleaving CEM and gradient descent steps can be used as an effective planner for model predictive control in the context of model based reinforcement learning.

Direct collocation approaches for control, address some of the ill-conditioning of shooting methods, and avoid backpropagating the model through time, by parameterizing the state and action sequences and optimizing both jointly. In this setting, ~\cite{hybrid} propose initializing the collocation optimization from a solution found by a genetic algorithm similar to CEM. However, they do not interleave the two optimizations and the collocation method requires parameterizing state trajectories with analytic functions such as splines.

  \vspace*{-0.2cm}
\section{Limitations and Future Works}
One of the main directions for future works is to investigate the implications of model-bias in the planning scheme. In MBRL, one of the primary issues leading to a suboptimal plan is that the planner exploits model bias of an imperfectly learned model~\cite{mbrl_survey}. So, for better planning, we also need to develop better strategies for learning the dynamics model itself. Some papers~\cite{pets,me_trpo} aim to learn a better model by maintaining an ensemble of neural network models. This helps model epistemic uncertainty, but an ensemble of networks for the dynamics is difficult to scale to image-based environments without introducing a huge computational burden during training.


Another effective direction for tackling model-bias is by learning dynamics models conditioned on some latent variables, instead of trying to learn a global dynamics model. A recent paper, DADS~\cite{dads} does this by conditioning the dynamics model on latent `skills' and the main idea is to learn smaller behavior-specific dynamics models instead of trying to learn a global dynamics model. The latent `skills' are basically an abstraction for the low-level action sequences that get executed in the environment. However, DADS does not leverage the latent abstractions for planning, it uses them only for learning the dynamics model. One potential extension of our approach would be to use such latent variable models for planning as well, by backpropagating gradients wrt the latent variables through the model, in order to update the low-level actions.

\section{Conclusion}
In this paper we investigate the problem of planning and optimization in model predictive control and in the context of model-based reinforcement learning. We address the scaling problems of the widely-used, but gradient-free, Cross-Entropy Method, which struggles as the dimensionality of the environment increases. This is an important issue as we scale these methods to real world control problems. On the other hand, gradient-descent-based planning is conveniently applicable to high-dimensional continuous control problems, especially since the learned dynamics models are typically parameterized by differentiable functions. We show that in environments with many local optima, pure gradient descent can fail to find an optimal solution, compared to CEM. Combining the strengths of the two approaches, we propose a simple method that interleaves CEM and gradient descent updates, and we show that this method scales to higher dimensions and performs at least as well as CEM on multi-extrema settings, while benefiting from the convergence speed of gradient-based optimization.   

\section*{Acknowledgement}
We thank the Natural Sciences and Engineering Research Council (NSERC) of Canada for funding support, and Vector Institute, Toronto, for computing infrastructure and support.

\bibstyle{neurips}
\bibliography{references}

\end{document}